\documentclass[sigconf]{acmart}
\renewcommand\footnotetextcopyrightpermission[1]{} 
\usepackage{booktabs} 
\usepackage{tabularx}
\usepackage{hyperref}
\usepackage{float}

\acmConference[]{}
\copyrightyear{}

\acmArticle{4}
\acmPrice{15.00}

\begin{document}
\title{Question Relevance in Visual Question Answering}
\author{Prakruthi Prabhakar}
\affiliation{%
  \institution{ML with Large Datasets (10-805) \\ Carnegie Mellon University}
  \city{Pittsburgh} 
  \state{PA} 
  \postcode{15217}
}
\email{prakrutp@andrew.cmu.edu}

\author{Nitish Kulkarni}
\affiliation{%
  \institution{ML with Large Datasets (10-805) \\ Carnegie Mellon University}
  \city{Pittsburgh}
  \state{PA}
  \postcode{15217}
}
\email{nitishkk@andrew.cmu.edu}

\author{Linghao Zhang}
\affiliation{%
  \institution{ML with Large Datasets (10-805) \\ Carnegie Mellon University}
  \city{Pittsburgh} 
  \state{PA} 
  \postcode{15217}
}
\email{linghaoz@andrew.cmu.edu}

\begin{abstract}
Free-form and open-ended Visual Question Answering systems solve the problem of providing an accurate natural language answer to a question pertaining to an image. Current VQA systems do not evaluate if the posed question is relevant to the input image and hence provide nonsensical answers when posed with irrelevant questions to an image. In this paper, we solve the problem of identifying the relevance of the posed question to an image. We address the problem as two sub-problems. We first identify if the question is visual or not. If the question is visual, we then determine if it's relevant to the image or not. 
For the second problem, we generate a large dataset from existing visual question answering datasets in order to enable the training of complex architectures and model the relevance of a visual question to an image. We also compare the results of our Long Short-Term Memory Recurrent Neural Network based models to Logistic Regression, XGBoost and multi-layer perceptron based approaches to the problem.
\end{abstract}

\maketitle

\section{Introduction}

The task of automatically answering questions in the context of visual information has gained prominence in the last few years. Being able to answer open-ended questions about an image is a challenging task, but one of great practical significance. For instance, visually impaired individuals might inquire about different aspects of an image in the form of free-form questions. However, when Visual Question Answering (VQA) systems are provided with irrelevant questions, they tend to provide nonsensical answers. VQA systems in real world scenarios are expected to be sophisticated to identify the relevance of the posed free-form questions to an input image, to answer them better. 

There are two aspects of relevance of a question to the input image:
\begin{enumerate}
\item Non-visual questions which do not require any input image to answer the question
\item False-premise questions which require an input image but do not pertain to the provided input image to answer the question\\
\end{enumerate}
In this project, we formulate the problem as follows:
\begin{quotation}
\noindent Given an image and a natural language question, identify if the question is relevant to the input image.
\end{quotation}

For visual versus non-visual question detection, we present the results of two approaches. The first approach is based on training a Logistic Regression model using unigrams, bigrams and trigrams of Part-of-Speech (POS) tags of the question. In the second approach, we use a Long Short-Term Memory (LSTM) Recurrent Neural Network trained on Part-of-Speech (POS) tags to capture the linguistic structure of questions. 

For the next sub-problem of identifying true versus false premise of a visual question to an image, we curate a much larger dataset compared to the existing datasets for the problem using variations in different existing data extraction methodologies. We also present the results of different models used for modeling the true versus false premise problem. In our initial approaches, we use Logistic Regression and XGBoost classifier using both visual and textual features to model the problem. We also explore several Long Short-Term Memory (LSTM) Recurrent Neural Network architectures and a multi-layer perceptron network to model the problem. Our code is available at \cite{github_link}. 

\section{Related work}

There have been significant advances in recent years on identifying the similarity between images and textual information. Text-based image retrieval \cite{liu2009boost} systems and visual semantic alignments in image captioning models \cite{karpathy2015deep}, \cite{fang2015captions} are some examples of efforts in that direction. While some systems do not answer when the input is ill-formed or likely to result in failure, some others try to find the most meaningful answer to such inputs. \cite{dodge2012detecting} tries separating visual text from non-visual text in image descriptions and use them for enhancing image captioning systems. These ideas can be used to boost the performance of visual question relevance task. 

Much of our work is based on the problem and approaches presented in \cite{ray2016question}. In this paper, the authors identify the two facets of question relevance for Visual Question Answering, i.e. categorizing the questions as visual versus non-visual questions, and then identifying if a question has a true premise for an image. The paper also provides several baselines for both  problems. For the visual versus non-visual classification, the authors propose a heuristic-based and an LSTM-based approach. And for the true versus false premise problem, there are three baselines based on entropy from VQA models, question-caption similarity and question-question similarity. 

For the problem to identifying false premise, \cite{mahendru2017promise} also makes significant contribution for extracting the premise from a question. The authors of this work also go on to create a well-curated, much larger and more class-balanced dataset for the true versus false premise based on the VQA dataset \cite{AntolALMBZP15}. In this paper, the concept of a premise in a question is explored in greater detail and a more diverse set of problems are addressed, such as - given an image and a question with a false premise, can we predict the premise in the question that cannot be answered by the image. However, the focus of our work is on exploring scalable algorithms and architectures for answering \textit{if} the question can be answered in the context of the image.

In the context of establishing semantic relationships between images and text, the work done in \cite{AkataMFS16} is also quite relevant, where the authors of this paper propose a holistic embedding technique for combining multiple and diverse language representations for better transfer of knowledge, by mapping visual and language parts into a common embedding space. Although our problem is supervised, we believe that such an embedding could help improve the identification of question relevance for an image.

\section{Dataset}

While Visual Question Answering has been a widely explored problem, question relevance for the same is relatively very new, owing to which there are no existing large-sized and diversely represented datasets. 




For the first task of detecting visual versus non-visual questions, we refer to the methodology used in \cite{ray2016question}. Since VQA 2.0 dataset \cite{goyal2016making} is now available, we use the training, validation and test questions for images from this dataset as visual questions. For non-visual questions, we use the philosophical and general knowledge questions provided by \cite{ray2016question}. Combining the two sources and eliminating duplicate questions, we have 
160,010 questions for training, 82,067 questions for validation and 148,927 questions for test datasets. The key limitations of this dataset are the class bias, and that the non-visual questions are not diversely representative of all possible non-visual questions.

The dataset for the second sub-problem of detecting questions with false premise is based on the VQA corpus \cite{AntolALMBZP15}. The data acquisition involves creating image-question pairs with true and false premises for the question. The questions in the VQA dataset can be assumed to have true premises, since they were manually generated.

To identify the questions with false premises, we choose to use the questions from the same dataset, but for other images. Here, we explore three approaches:

\begin{enumerate}

\item \textbf{Question Similarity} \\[-10pt]

In this approach, for every image, we use the set of true-premise questions from the VQA dataset \cite{AntolALMBZP15}, and extract $k$ least similar questions from the set. As similarity measure, we try using doc2vec similarity and word2vec similarity for keywords in a question (nouns, verbs and adjectives). \\[-10pt]

\item \textbf{Visual True vs False Premise Questions (VTFQ) Dataset} \\[-10pt]

In this approach, we use the dataset presented in \cite{ray2016question}, where the methodology is similar to the one described for question similarity, but instead of using a question similarity measure, the authors sample random questions from the set, and have Amazon Mechanical Turk (AMT) workers annotate the questions as relevant or irrelevant to the corresponding images. The VTFQ dataset consists of 10,793 question-image pairs with 1,500 unique images of which 79\% of the pairs have false premise. \\[-10pt]

\item \textbf{Question Relevance Prediction and Explanation (QRPE) Dataset} \\[-10pt]

This dataset is presented by \cite{mahendru2017promise}, with a methodology that uses VQA 1.0 questions, COCO images and Visual Genome annotations. The approach first involves a premise-extraction for a given question, where a question premise is defined as a semantic tuple comprising of objects, objects and their attributes or objects and their relationships in an image. Then, for a given question-image pair in the VQA dataset, a set of all images is created which has exactly one premise as false for the question, which are referred as negative images. To make the false-premise detection problem challenging, the negative images that are most similar to the positive image (i.e. image with true premise for the question) are chosen. The QRPE dataset consists of 53,911 tuples of the form $(I^+, Q, P, I^-)$, where $(I^+, Q)$ is a pair of positive image and true premise question, $I^-$ is the negative image with $P$ as the premise in $Q$ that is false for $I^-$. While a single QI pair for true and false premises is extracted using this approach.

\end{enumerate}

The problem with the first two approaches is that using random or least similar questions for an image would make the problem of false-premise detection much easier than the case where the a single premise of the question were to be false in the context of the image or if the negative images are very similar to the positive images. While this is addressed in QRPE dataset, it has only 53,911 positive and negative image pairs, generated from the VQA dataset that is much larger to begin with. This is largely because of the constraints imposed in the dataset construction such as restricting the questions and objects to be of specific categories. The constraints are placed in order to minimize the noisy samples since the data is generated heuristically. Since the dataset is too small, most of the current state-of-the-art models use pre-trained VQA or image captioning models trained on COCO dataset.

We hypothesize that a much larger dataset with only a marginal reduction in robustness would help build a more effective model for classifying true vs false premise, since a large dataset would enable end-to-end training of deep architectures to optimize the classification performance for this task. To test this hypothesis, we present an extended QRPE dataset that is built by making some modifications to the methodology in \cite{mahendru2017promise}.

The first difference is that we use VQA 2.0 \cite{goyal2016making}, which has twice as many image-question pairs and also contains complementary image-question pairs. In addition, we relax some constraints on question types (while regulating robustness of the dataset) as well as increase the number of negative images generated for each question from 1 to 10. Lastly, we use all the image-question pairs in VQA 2.0 as true-premise pairs, including the ones that do not have negative images. This is because the goal of our dataset is to have a large dataset with good representation of both the classes, and not to generate $(I^+, Q, P, I^-)$ tuples. \\

\noindent \textbf{Dataset Construction:}

\noindent
Since the true premise instances are directly derived from the VQA 2.0 dataset, the key challenge lies in generating a mapping of images to questions with false premises. To ensure that the irrelevance of the question does not become trivial, we use only the top 10 most similar images to the positive image for the question. To compute the similarity of images, cosine similarity of VGG 16 image features\cite{SimonyanZ14a} is used, since it has been pre-trained on 1.3M images. To identify the set of image-question pairs with false premise, two kinds of premises are considered:
\begin{enumerate}
\item \textbf{First Order Premises}:
These are the existential premises that represent only the presence of objects, such as \textit{cat}, \textit{dog} etc. To generate image-question pairs with negative first order premises, we use the class annotations from the COCO dataset and check for their presence in the question. 

\item \textbf{Second Order Premises}:
These contain the objects and their attributes, such as \textit{small cat}, \textit{black dog} etc. For a false second order premise, we consider those images with true first order premise i.e. with the object present in the image, and look for the opposite attribute. To get the premises for a given image, we construct scene graphs by using the semantic tuple extraction pipeline used in the SPICE metric \cite{AndersonFJG16}, which is an image captioning metric. We compare these premises with the Visual Genome scene graphs for the COCO images, and use the images whose attribute is an antonym of the attribute in the question.

\end{enumerate}

Table \ref{table:dataset} shows the data characteristics of the dataset created using this methodology and the applied modifications.

\begin{figure}[t]
\begin{center}
  \includegraphics[width=0.9\linewidth]{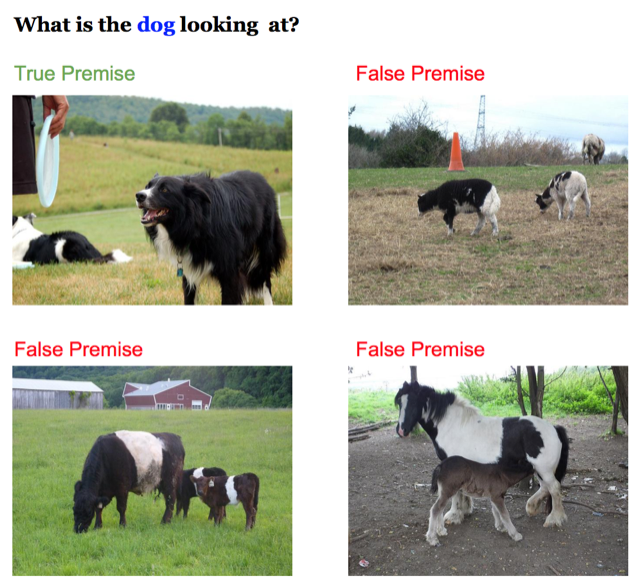}
\end{center}
   \caption{Positive (top left) and negative (top right, bottom left and bottom right) images based on first-order premises for a sample question (top).}
\label{fig:first_order}
\end{figure}

\begin{figure}[t]
\begin{center}
  \includegraphics[width=0.9\linewidth]{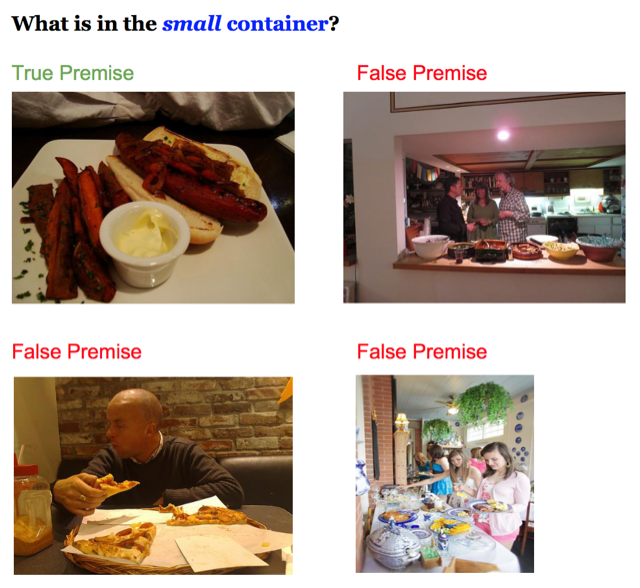}
\end{center}
   \caption{Positive (top left) and negative (top right, bottom left and bottom right) images based on second-order premises for a sample question (top).}
\label{fig:second_order}
\end{figure}

Some negative images generated for false first and second order premises are illustrated in Figures \ref{fig:first_order} and \ref{fig:second_order}. We can notice how the objects in the negative images for the first order premise are different but look very similar to the object in the positive image (in this case, the dog). On the other hand, for the second order premise, the object (container) is the same in the negative images, but the attribute is different (large vs small). Thus, it is much harder to identify false second order premises than false first order premises. In some cases, it is not very obvious if the premise is false, for example, in this case, it may not be clear if the container should be considered large or small. This is why we focus more on classifying the first order pairs, as they are more definitive.

\begin{table}[t]
	\centering
    \begin{tabular}{c|c|c|c} \toprule
    			& Total & Relevant & Non-relevant \\ \hline
        Total 	& 3,697,728 & 1,551,009 & 2,146,719 \\ \hline
        First order & 2,671,037 & 658,111 & 2,012,926 \\ \hline
        Second order & 1,026,691 & 892,898 & 133,793 \\ \bottomrule
    \end{tabular}
    \caption{Extended QRPE dataset characteristics for true vs false premise}
    \label{table:dataset}
\end{table}

\section{Approach}

Among the different models experimented with, we first present the ones for visual-vs-non-visual question detection, followed by the models used for true-vs-false premise question relevance problem.

\subsection{Visual vs. Non-visual Question Detection}
We identify that non-visual questions have different linguistic structure than visual questions. For example, non-visual questions such as "Name the national rugby team of Argentina." or "Who is the president of Zimbabwe?" often have differences in structure in comparison to visual questions such as "Is this truck yellow?" and "What color are the giraffes?". Hence, we use Spacy \cite{spacy} to process all questions to obtain Part-of-Speech (POS) tags as features. We compare two models, a Logistic Regression model versus an LSTM-RNN based approach.

\begin{enumerate}
\item \textbf{Logistic Regression} We trained a Logistic Regression model using POS tags of the questions as features. We also experimented with larger feature sets using bigrams and trigrams of POS tags. We implemented a scalable streaming version of logistic regression for training. We assume that the validation and test datasets fit in memory for this problem. 

\item \textbf{LSTM} We also trained an LSTM model using the architecture from \cite{ray2016question} for modeling visual vs. non-visual question detection. This architecture is shown in Figure \ref{fig:LSTM0}. This model uses a dimensionality of 100 for hidden vectors and POS tags of the words in the question as the sequence input to the LSTM. We also experimented with alternate architectures with varying dimensionality of the hidden vectors as well as POS-tag embeddings.
\begin{figure}[t]
\begin{center}
  \includegraphics[width=0.8\linewidth]{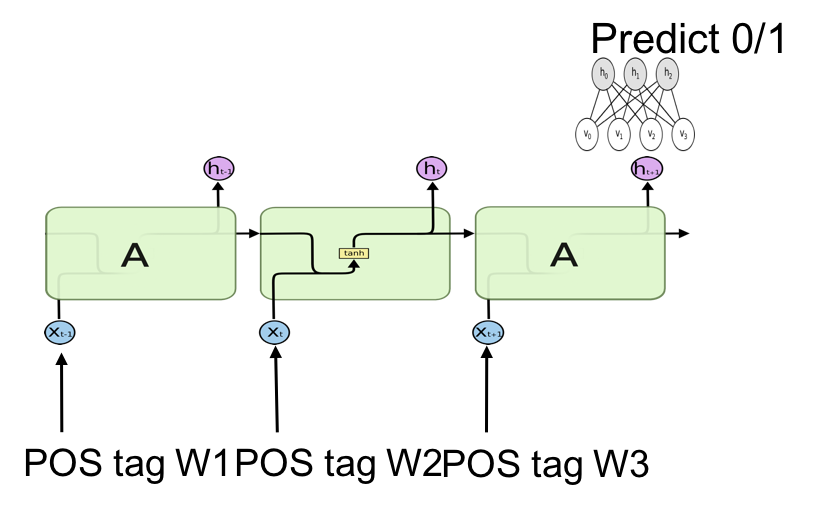}
\end{center}
   \caption{LSTM based architecture for modeling visual versus non-visual question detection. The POS tag of every word is input to the LSTM layer, which is used to predict if the question is visual or not.}
\label{fig:LSTM0}
\end{figure}
\end{enumerate}

\subsection{True vs. False Premise Question Relevance Detection}

While visual versus non-visual question detection depended only on the posed textual question, the true versus false premise question relevance requires joint modeling of image and the question. We obtain visual and textual representations to model them together. For all our models, we use a pre-trained VGGNet \cite{simonyan2014very} convolutional neural network and obtain the fully connected seventh layer of the network output as visual features for the images.

\begin{enumerate}
\item \textbf{Logistic Regression} Our first approach to modeling the problem of true versus false premise question relevance was to use a scalable Logistic Regression classifier combining visual and textual features for the problem. For the visual features, we used Principal Component Analysis to reduce the 4096 dimensional FC7 features of an image obtained from a pre-trained VGGNet to 300 dimensions. For the textual features, we trained a FastText model \cite{bojanowski2016enriching} on all the VQA questions to obtain average word embeddings for the questions as features. We combine these representations to learn a model to classify if the question is relevant to the image or not.
\item \textbf{XGBoost Gradient Boosting Classifier} We use the same visual and textual feature representations described above for training an XGBoost Gradient Boosting Classifier \cite{chen2016xgboost}. We use disk-based implementation of XGBoost to accommodate the large training dataset size.
\item \textbf{Multi-Layer Perceptron} We also explore a multi-layer perceptron using 4096 image features concatenated with 300 dimensional Glove embedding features of the words in the question as input. We use two hidden layers with 5000 and 500 hidden units respectively. The output layer has one unit modeling the probability of question relevance using a binary cross-entropy loss.
\item \textbf{LSTM} We also train different variants of LSTM architecture for jointly modeling image and textual inputs. These architectures are described briefly below. All our architecture variants are named as RelNet (Relevance Net for Question Relevance) for reference. In all architectures, the question is input the the LSTM using an Embedding layer. We initialize the embedding layer using 300 dimensional Glove embeddings \cite{pennington2014glove}.
\begin{enumerate}
\item \textbf{RelNet1} This architecture is shown in Figure \ref{fig:LSTM1}. In this network, we used Principal Component Analysis to reduce the 4096 dimensional FC7 features of an image obtained from a pre-trained VGGNet to 300 dimensions. We use an LSTM layer to model the input question. The output of this LSTM layer is concatenated with the image features at every time step and then fed to another LSTM layer. This LSTM layer is trained to model the probability of the question being relevant to the image or not using a binary cross entropy loss.
\begin{figure}[t]
\begin{center}
  \includegraphics[width=0.8\linewidth]{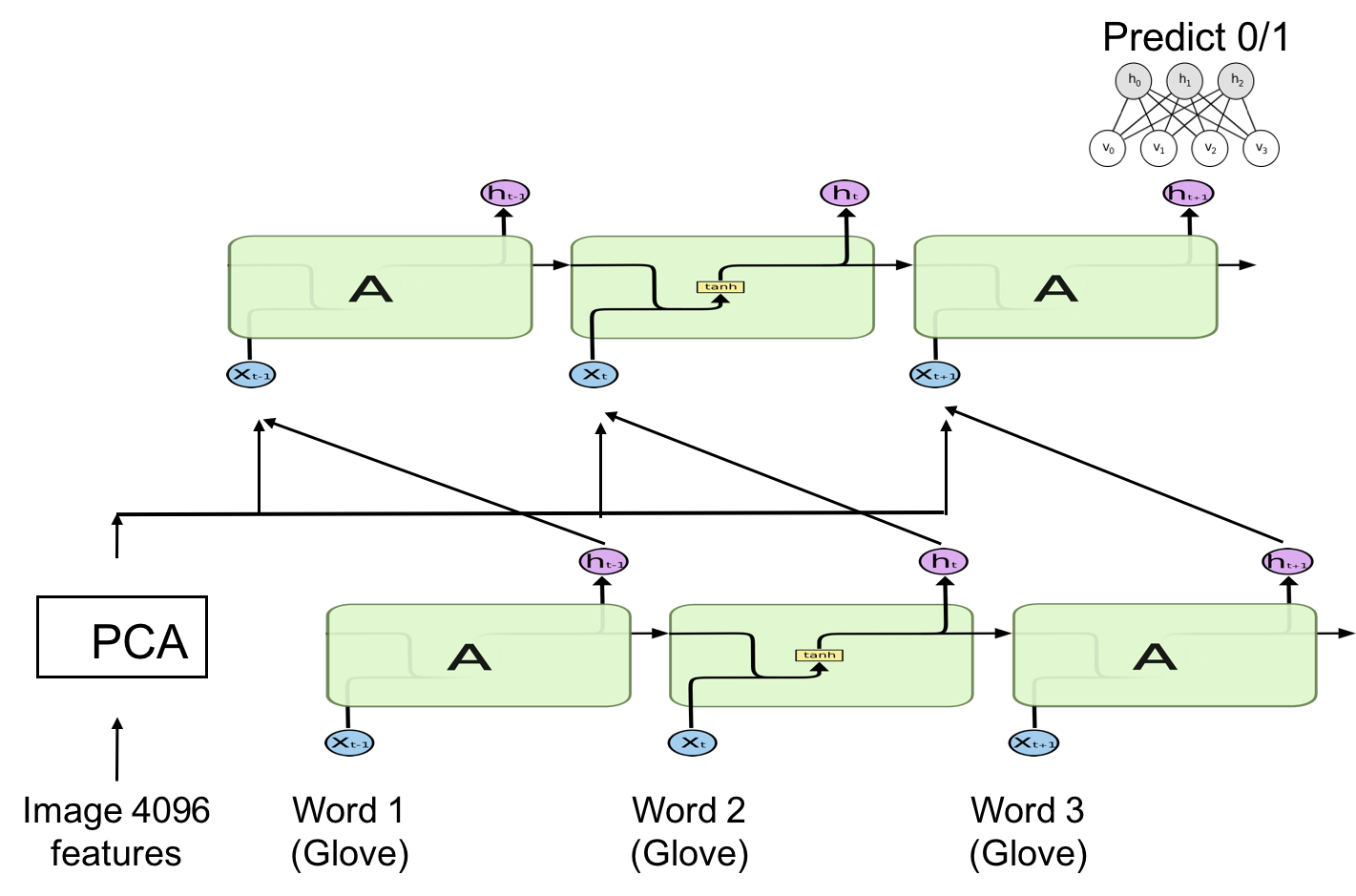}
\end{center}
   \caption{RelNet1: Image features after PCA is input to an LSTM layer at every time step. The Question is modeled using another LSTM layer whose output is also input to the final LSTM layer.}
\label{fig:LSTM1}
\end{figure}
\item \textbf{RelNet2} This architecture is shown in Figure \ref{fig:LSTM2}. In this network, we modify the architecture of RelNet1 by training an embedding layer to reduce the 4096 dimensional FC7 features of an image to 300 dimensions, instead of using Principal Component Analysis.
\begin{figure}[t]
\begin{center}
  \includegraphics[width=0.8\linewidth]{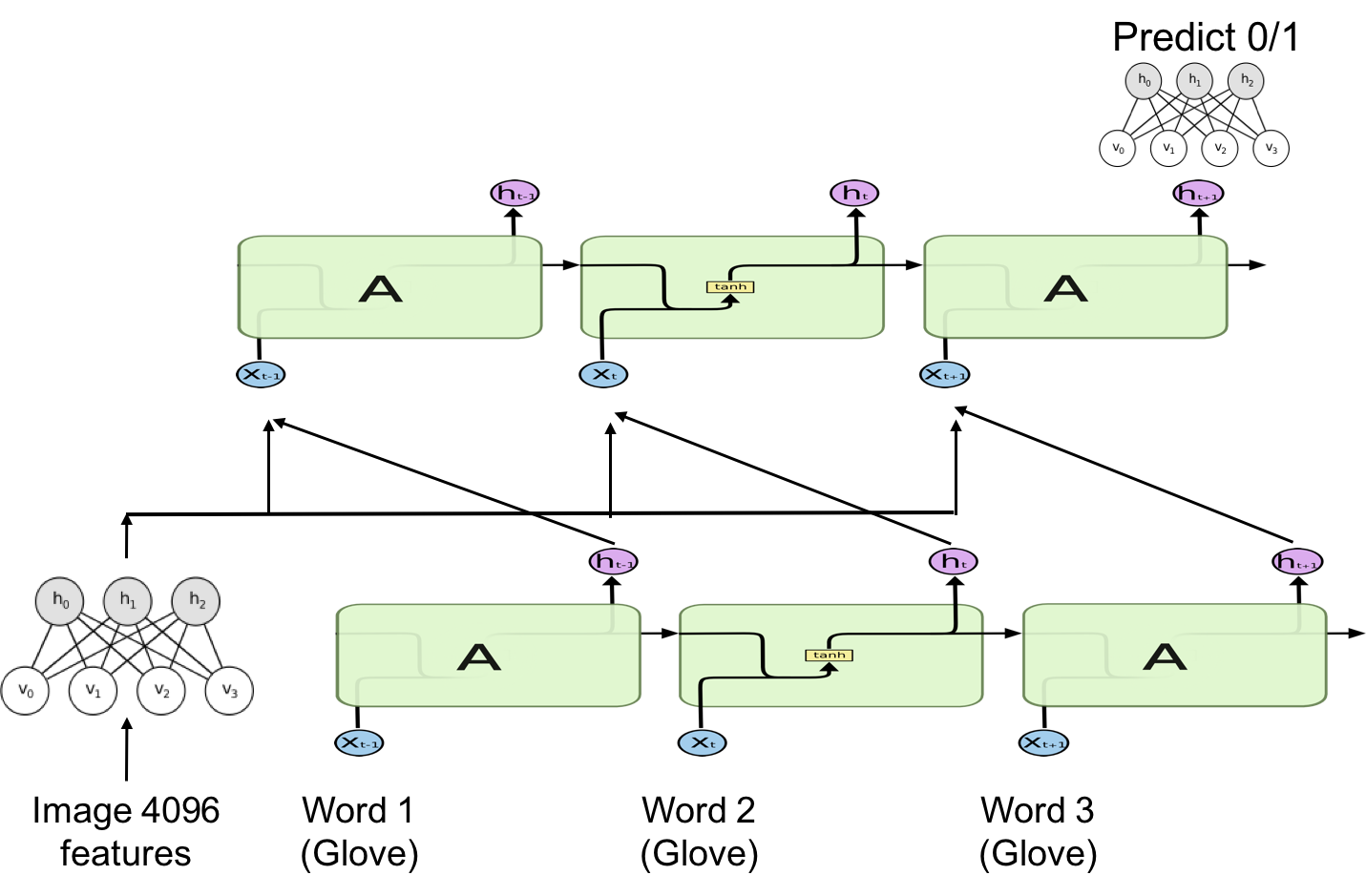}
\end{center}
   \caption{RelNet2: Image features are input to an LSTM layer at every time step by passing through a linear embedding layer. The Question is modeled using another LSTM layer whose output is also input to the final LSTM layer.}
\label{fig:LSTM2}
\end{figure}
\item \textbf{RelNet3} This architecture is shown in Figure \ref{fig:LSTM3}. In this network, we modify the architecture of RelNet2. We do not concatenate the image features with every time step output of the first LSTM layer. Instead, we input the image features at the first time step only to the final LSTM layer. The output of the language LSTM layer is fed to the final LSTM layer from the second time step.
\begin{figure}[t]
\begin{center}
  \includegraphics[width=0.8\linewidth]{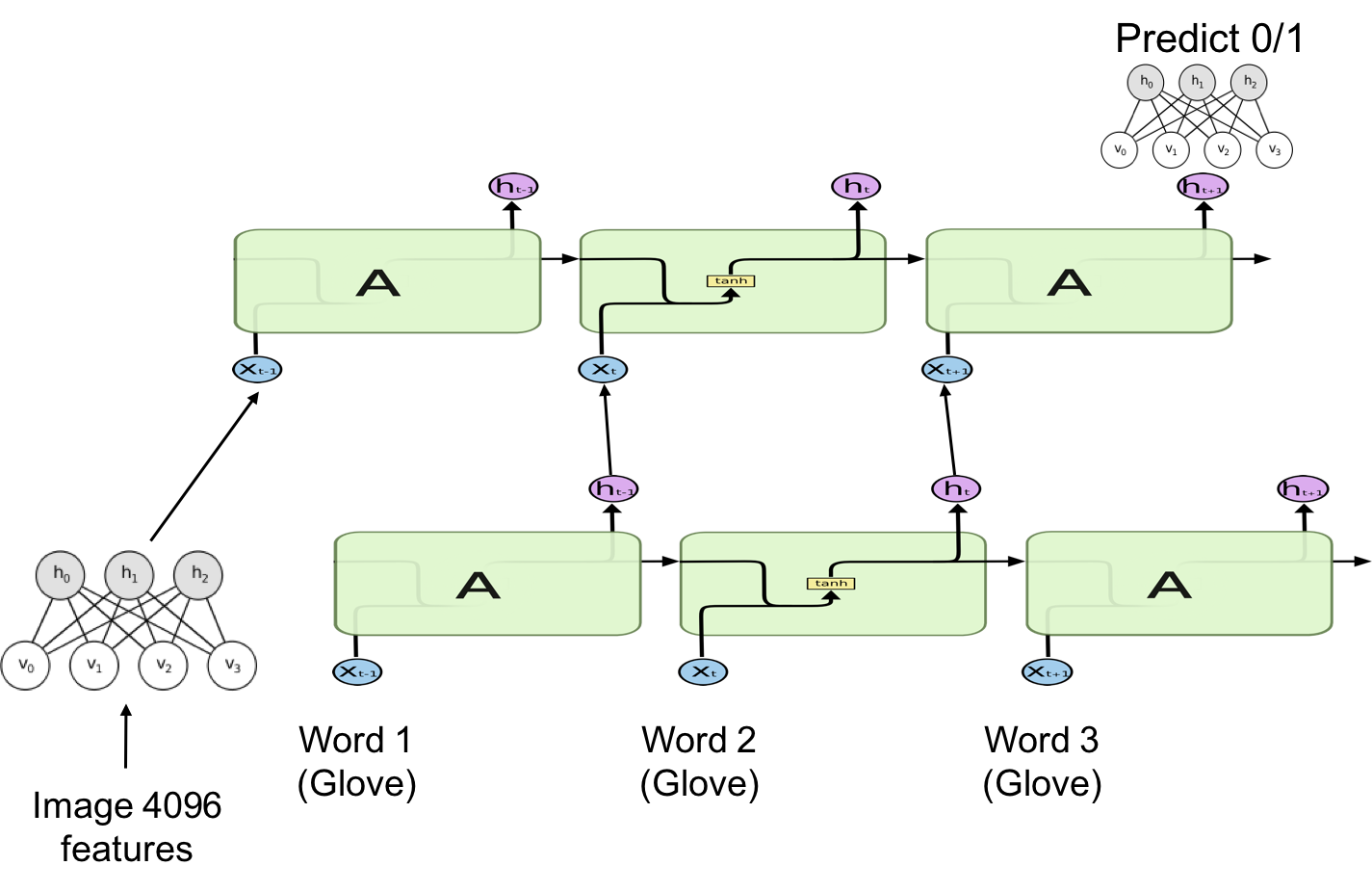}
\end{center}
   \caption{RelNet3: Image features are input to an LSTM layer only at the first time step by passing through a linear embedding layer. The Question is modeled using another LSTM layer whose output is input to the final LSTM layer from the second time step onwards.}
\label{fig:LSTM3}
\end{figure}
\item \textbf{RelNet4} This architecture is shown in Figure \ref{fig:LSTM4}. In this network, we do not have the first LSTM layer to model the input question. We directly input the question to the final LSTM layer from the second time step using an Embedding layer.
\begin{figure}[t]
\begin{center}
  \includegraphics[width=0.8\linewidth]{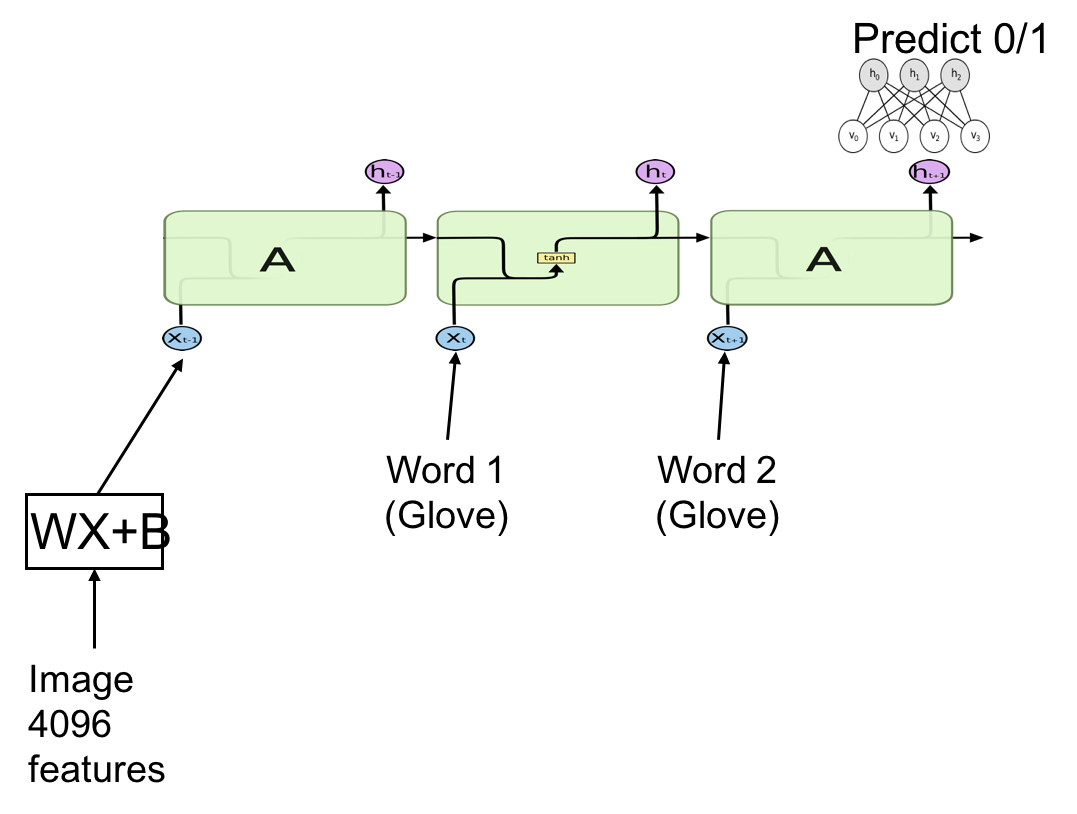}
\end{center}
   \caption{RelNet4: Image features are input to an LSTM layer only at the first time step by passing through a linear embedding layer. The Question is input to the LSTM layer from the second time step onwards using an Embedding layer.}
\label{fig:LSTM4}
\end{figure}
\end{enumerate}
\end{enumerate}

\nocite{colah_blog}

\section{Experiments and results}
We compare our models with several baseline approaches. We model LSTM based methods on a NVIDIA Tesla K80 GPU. The following sections presents the results for various models for both sub-problems.

\subsection{Visual vs. Non-Visual Question Detection}
The results for visual vs. non-visual models are presented in Table \ref{tab:results_visual_lr} and Table \ref{tab:results_visual}. Since there is a class imbalance problem in the datasets, we report the average per-class (i.e., normalized) metrics for all approaches.

Table \ref{tab:results_visual_lr} compares the results for the three models of Logistic Regression, using unigrams, bigrams and trigrams of POS tags as features. As can be observed from the table, addition of bigrams of POS tags as features helped improve the precision and recall of both classes significantly in comparison to using only unigrams of POS tags as features. However, using trigrams as additional features didn't give significant improvement in the metrics. Additionally, using trigrams as features increases computational time exponentially. Hence, we use unigram and bigram based logistic regression to compare with results from LSTM model. 

We trained several models of LSTM with varying embedding and hidden vector dimensionality using a NVIDIA Tesla K80 GPU. We observed that all models performed similarly across different metrics. Hence, we provide the results of replicating the model provided in \cite{ray2016question}. Since we use VQA 2.0 dataset \cite{goyal2016making} and different set of POS tags from Spacy, we identify that the results are different from the original paper for the same model.

\begin{table}
  \caption{Results of visual versus non-visual question detection using different POS features in Logistic Regression. Here, Uni represents using unigram POS tags as features, Uni+Bi represents using unigram as well as bigram POS tags as features, Uni+Bi+Tri represents using unigram, bigram and trigram as features.}
  \label{tab:results_visual_lr}
  \begin{tabular}{lccc}
    \toprule
    Metric&Uni&Uni+Bi&Uni+Bi+Tri\\
    \midrule
    Precision (Visual class)&0.9990&0.9996&0.9997\\
    Recall (Visual class)&0.9980&0.9997&0.9993\\
    Precision (Non-visual class)&0.8193&0.9690&0.9354\\
    Recall (Non-visual class)&0.9049&0.9638&0.9705\\
  \bottomrule
\end{tabular}
\end{table}
\begin{table}
  \caption{Comparison of results of Logistic Regression and LSTM based models for visual versus non-visual question detection}
  \label{tab:results_visual}
  \begin{tabular}{lcc}
    \toprule
    Metric&Logistic Regression&LSTM\\
    \midrule
    Precision (Visual class)&0.9996&0.9995\\
    Recall (Visual class)&0.9997&1.0\\
    Precision (Non-visual class)&0.9690&1.0\\
    Recall (Non-visual class)&0.9638&0.9511\\
  \bottomrule
\end{tabular}
\end{table}

\subsection{True vs. False Premise Question Relevance Detection}
Table \ref{tab:results_premise} compares the accuracy metric for various models on the three datasets. The two baseline models presented in the table are obtained from \cite{mahendru2017promise}. VQA-Bin baseline is modeled using a pre-trained deeper LSTM VQA architecture with fine-tuning for the binary question relevance detection task. QPC-Sim baseline uses a pre-trained image captioning model to automatically provide natural language image descriptions
and identifies question relevance based on a learned similarity between the question, the premise and the generated image caption. Since the baselines were not trained on our dataset, we present the baseline results for the QRPE dataset's first order and second order image question pairs. All our models are trained on the generated first order train dataset and tested on the generated first order test dataset, second order test dataset and QRPE test dataset.

\begin{table}
  \caption{Results of true premise versus false premise question relevance detection using different models.}
  \label{tab:results_premise}
  \begin{tabularx}{\linewidth}{|l|X|X|X|}
    \toprule
    Model&First-order dataset&Second-order dataset&QRPE dataset\\
    \midrule
    VQA-Bin (Baseline 1)&0.6736&0.53&0.665\\
    QPC-Sim (Baseline 2)&0.7667&0.5595& \textbf{0.7531}\\
    Logistic Regression&0.6784&0.4293&0.5746\\
    XGBoost&0.8622&0.7466&0.6206\\
    MLP&0.8886&0.7507&0.6083\\
    RelNet1& \textbf{0.8889} & - & 0.6573 \\
    RelNet2&0.8825&-&0.6606\\
    RelNet3&0.8878&0.7893&0.6564\\
    RelNet4&0.8848&\textbf{0.7945}&0.6623\\
  \bottomrule
\end{tabularx}
\end{table}

All our models are trained end-to-end without relying on other tasks like image captioning and visual question answering. By using a larger dataset, we are able to obtain good performance on the QRPE dataset by directly modeling question relevance. Usage of pre-trained models like VQA and image captioning is limiting because of the need to have relevant datasets for those tasks which have similar images as the question relevance task. These models have inherent errors in their tasks and also induce additional errors for question relevance datasets which don't share the same images. Hence, by generating a larger dataset, we were able to directly model the question relevance detection task reasonably well. 

\section{Analysis}

For the first task of visual vs non-visual question detection, from table \ref{tab:results_visual} we can observe that logistic regression performs better than LSTM based approach on some metrics and performs comparatively on others. However, we provide a scalable streaming implementation of logistic regression, which takes significantly less time for training in comparison to LSTM. It can also be inferred from the high precision and recall that this is a much simpler problem compared to the second sub-problem. One of the possible reasons for this, we believe, is that the data for non-visual questions is not well represented (since most of them are general-knowledge or philosophical questions) and the complex models are undesirably learning the ad-hoc attributes of the two classes. A possible future work in this regard would be to collate a richer set of datasets for non-visual questions from multiple sources.

For the true versus false premise detection, it can be observed from the results in table \ref{tab:results_premise} that our models perform quite well compared to the baselines on the extended first order and second order dataset. Among the LSTM architectures, RelNet1 and RelNet4 performed well on the test dataset. RelNet1 has the PCA dimensionality reduced information in it's image features, thereby eliminating the need to learn a rich image representation from the training data. RelNet4 is a much simpler model in terms of number of parameters and network layers for the model to learn. Hence, we believe that these two networks provided good results on the test datasets.

From a computational perspective, all the RelNets and the multi-layer perceptron model took 12-17 minutes per epoch and we trained each of the models for 20-30 epochs. Since the generated first-order dataset is large, we used a training data generator to generate images batch by batch. We use Keras to code all our models. For the XGBoost model, we use the open source disk-based implementation of XGBoost \cite{chen2016xgboost}. The Logistic Regression model is a streaming implementation which avoids loading the entire training data into memory.

In order to test the generalization of our models as well as the quality of our dataset, we have also tested our models on the QRPE dataset, and found that the performance is as good as some of the baselines (VQA-bin), but not as good as the best performing models (QPC-Sim). This can be attributed to two key reasons. The first one is that the models on QRPE dataset are a lot more complex, since they use pre-trained VQA and image captioning models. Secondly, the construction of the dataset may itself introduce a bias in the model. A possible way to circumvent this bias is to use a combination of differently constructed datasets (like QRPE and VTFQ) as validation sets and minimize the generalization loss while training. Lastly, from the consistent under-performance of our models on the QRPE dataset compared to our test dataset, we can also infer that the classification task on QRPE dataset is more challenging, since we have loosen the constraints for filtering question types while constructing the extended dataset.

\section{Conclusion And Future Work}

In this project, we attempt the problem of identifying relevance of posed questions to visual question answering systems by exploring several approaches that yield similar or better results (by virtue of larger training data). For the first sub-problem of identifying visual versus non-visual question detection, we provide a time-efficient and scalable implementation of logistic regression. This approach provides comparative or better results on all metrics in comparison to strong baselines set by LSTM based approaches. To solve the second sub-problem of true versus false premise using end-to-end classifiers, we propose an extended QRPE dataset using a modified QRPE data-generation pipeline and VQA 2.0, consisting of over 3.6M image-question pairs. We then experiment with multiple families of models for classifying true versus false premise such as XGBoost, Logistic Regression and LSTM-based RNN models. Each of these approaches have been trained end-to-end on the generated first order dataset, thereby avoiding using pre-trained VQA and image captioning models. While the models performed better than the baselines on our dataset, it did not outperform the state-of-the-art model (QPC-Sim) on the QRPE dataset, which we believe is because our models have much simpler architectures compared to the pre-trained models that QPC-Sim uses.

There are many directions for future work in this area. For the dataset of true versus false premise detection, third order false premises (which include the relationships between objects in the image) could also be included. Given the larger data, many deep generative models can be trained to outperform the pre-trained models. As for features, we have considered only CNN features for images and word embeddings for words, but many different possibilities of imaged and words can be explored, with an option of training the language model and CNN specifically for this task. A natural extension to the problem of question relevance using premises is to explain why the question is not relevant, i.e. which premises are false and what additional information is required to answer the question.

With greater capabilities in identifying and commenting on the relevance of the questions for images, visual question answering would have a much larger applicability in a practical setting.

\bibliographystyle{ACM-Reference-Format}
\bibliography{bibliography} 

\end{document}